\documentclass[runningheads]{llncs}

\usepackage{eccv}

\usepackage{eccvabbrv}

\usepackage{graphicx}
\usepackage{booktabs}

\usepackage[accsupp]{axessibility}  

\usepackage{hyperref}

\usepackage{orcidlink}

\usepackage[dvipsnames]{xcolor}

\newcommand{\rone}[1]{{\bf \color{Violet} WoTQ}}
\newcommand{\rtwo}[1]{{\bf \color{Maroon} ijV4}}
\newcommand{\rthree}[1]{{\bf \color{teal} LtLd}}

\graphicspath{{figures/}}

\newcommand{\Eq}[1]{Eq.~(\ref{eq:#1})}
\newcommand{\eq}[1]{\Eq{#1}}

\newcommand{\fig}[1]{Fig.~\ref{fig:#1}}
\newcommand{\tab}[1]{Table.~\ref{tab:#1}}
\newcommand{\tabl}[1]{Tab.~\ref{table:#1}}

\newcommand{\sect}[1]{Section~\ref{sec:#1}}

\usepackage{graphicx}
\usepackage{amsmath}
\usepackage{amssymb}
\usepackage{booktabs}
\usepackage{comment}

\usepackage{mathtools}

\usepackage{color, soul}

\usepackage{multirow}

\usepackage{pifont}
\newcommand{\cmark}{\ding{51}} 
\newcommand{\xmark}{\ding{55}} 

\begin{document}

\title{MSD: A Benchmark Dataset for Floor Plan Generation of Building Complexes} 

\titlerunning{MSD: A Dataset for Floor Plan Generation of Building Complexes}

\author{
    Casper van Engelenburg
    \inst{1}
    \orcidlink{0000-0002-9705-9644}
    \and
    Fatemeh Mostafavi
    \inst{1}
    \orcidlink{0000-0002-8047-2168}
    \and
    Emanuel Kuhn
    \inst{1}
    \and
    Yuntae Jeon
    \inst{2}
    \orcidlink{0000-0002-1777-5297}
    \and
    Michael Franzen
    \inst{3}
    \orcidlink{0000-0002-1740-2685}
    \and
    Matthias Standfest
    \inst{3}
    \orcidlink{0000-0001-9592-6217}
    \and
    Jan van Gemert
    \inst{1}
    \orcidlink{0000-0002-3913-2786}
    \and
    Seyran Khademi
    \inst{1}
    \orcidlink{0000-0003-4623-3689}
}


\authorrunning{C. van Engelenburg et al.}

\institute{
    Delft University of Technology, The Netherlands
    \\ 
    Sungkyunkwan University, Korea
    \\
    Independent Researcher
    }

\maketitle

\begin{abstract}

Diverse and realistic floor plan data are essential for the development of useful computer-aided methods in architectural design. 
Today's large-scale floor plan datasets predominantly feature simple floor plan layouts, typically representing single-apartment dwellings only.
To compensate for the mismatch between current datasets and the real world, we develop \textbf{Modified Swiss Dwellings} (MSD) -- the first large-scale floor plan dataset that contains a significant share of layouts of multi-apartment dwellings. 
MSD features over 5.3K floor plans of medium- to large-scale building complexes, covering over 18.9K distinct apartments.
We validate that existing approaches for floor plan generation, while effective in simpler scenarios, cannot yet seamlessly address the challenges posed by MSD. 
Our benchmark calls for new research in floor plan machine understanding. 
Code and data are open.


\end{abstract}
\section{Introduction}

A floor plan is a 2D horizontal projection of a building’s floor, effectively conveying the layout of its inherent spatial components, such as areas, doors, and walls.
Developing floor plans is a primary task in architectural design, and is a time-consuming and expensive operation -- it is an informal optimization of multi-variable space functionality, concerning various constraints (\eg, environmental context, regulations, budget).

\begin{figure*}[t]
    \centering
    \includegraphics[width=1\textwidth]{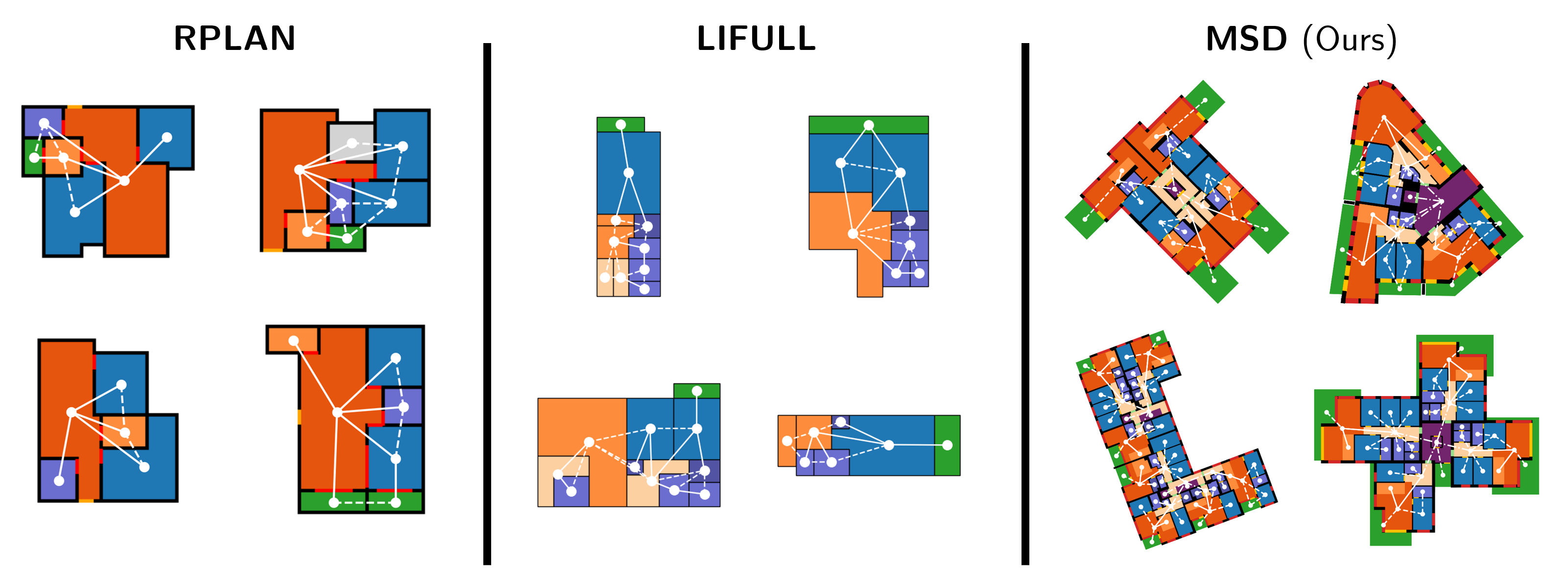}
    \small
    \caption{
        \textbf{MSD compared to RPLAN~\cite{wu_data-driven_2019} and LIFULL~\cite{lifull_co_ltd_lifull_nodate}.} 
        Rooms are colored on function (\eg, blue for "bedroom"). 
        The functional diagrams (represented as graphs) are drawn on top of the floor plans. 
        MSD (right) significantly differs from RPLAN (left) and LIFULL (middle), as it contains more complex and realistic floor plans. 
    }
    \label{fig:motivational}
\end{figure*}

Recent advancements in deep learning and the accessibility of large-scale floor plan datasets~\cite{kalervo_cubicasa5k_2019, wu_data-driven_2019}, led to a large number of techniques for automatically generating floor plans~\cite{wu_data-driven_2019, nauata_house-gan_2020, shabani_housediffusion_2023, tang_graph_2024}. 
The main focus of the current works has been on simple floor plans, mostly of small-scale and single-apartment dwellings. However, the majority of real-world dwellings are more complex, especially those that consist of multiple apartments.

Floor plans of multi-apartment building complexes are very different from single-apartment floor plans. 
Not only are there an order of magnitude more areas that need to be arranged, but the connectivity \textit{between} apartments plays an essential role as well. 
Moreover, there are structural constraints on the floor plan design (\eg, staircases, load-bearing walls) that need to remain intact while arranging the space.

To train and evaluate realistic models, we curate a new floor plan dataset, called Modified Swiss Dwellings (MSD), that consists of a large number of complex floor plans.
MSD includes precise area annotations, graph attributes, and the essential structural components of the building.
Following~\cite{hu_graph2plan_2020}, we define the floor plan generation task as one that is constrained on the functional diagram (represented as a graph) and the necessary structure of the building (represented as a binary image). 
To benchmark the complexity of MSD, we develop two baseline methods.
The first method modifies \textit{HouseDiffusion}~\cite{shabani_housediffusion_2023} by including a wall cross attention module and by integrating it with a graph attention network (GAT)~\cite{velickovic_graph_2018}. 
The second is a segmentation-based approach, integrating a U-Net~\cite{ronneberger_u-net_2015} and a graph convolutional network (GCN)~\cite{kipf_semi-supervised_2017}. 

Our results reveal a significant drop in performance when the two methods are trained and tested on MSD compared to the performance on simpler floor plan data.
The increased complexity that MSD brings underscores the need to reassess the current methods for floor plan generation. Our contributions are summarized as follows:

\begin{itemize}
    \item[$\bullet$] 
    We develop MSD -- a benchmark dataset of floor plans of building complexes. MSD contains 5,372 annotated floor plan images of medium- to large-scale single- to multi-apartment building complexes, including precise geometrical and topological attributes. 
    
    \item[$\bullet$] 
    We benchmark two state-of-the-art frameworks for floor plan generation to validate the complexity of floor plan generation of building complexes on MSD.  
    
    \item[$\bullet$] 
    Our evaluations of generated floor plans for the two baseline methods reveal that the floor plan generation of building complexes is a very challenging task and invites researchers to rethink current methods.
\end{itemize}

\section{Related work}\label{sec:review}

\paragraph{\textbf{Floor plan datasets}.} 
Floor plan datasets are used for retrieval~\cite{sharma_daniel_2017}, reconstruction and structural reasoning \cite{de_las_heras_cvc-fp_2015, liu_raster--vector_2017, kalervo_cubicasa5k_2019, surikov_floor_2020, lu_data-driven_2021}, architectural symbol spotting and wall detection \cite{goyal_bridge_2019, swaileh_versailles-fp_2021, dodge_parsing_2017}, and floor plan generation~\cite{wu_data-driven_2019, lifull_co_ltd_lifull_nodate}.  
For further comparison, we only consider publicly available datasets used in floor plan generation: RPLAN~\cite{wu_data-driven_2019} and LIFULL~\cite{lifull_co_ltd_lifull_nodate} (the part that is publicly available).
RPLAN and LIFULL contain 80K+ and 177K+ floor plans of, resp., Japanese and Asian houses. 
While large in scale, RPLAN and LIFULL have a significant number of shortcomings. 
First, both datasets only cover single-apartment dwellings with a limited number of areas.
Second, floor plans in RPLAN and LIFULL are axis-aligned and entirely Manhattan-shaped layouts, which is at odds with realistic dwellings, which typically contain a significant number of more irregularly shaped rooms. 
Third, RPLAN and LIFULL do not provide compass orientation, while the direction of the sun is a critical feature in environmental design~\cite{mostafavi_energy_2021}.
In our work, we gather and develop MSD -- a big collection of floor plans that addresses the above-mentioned limitations. Specifically, MSD comprises single- \textit{and} multi-apartment dwellings of which a significant part contains irregularly shaped areas as well as building boundaries.

\paragraph{\textbf{Automated floor plan generation}.}
The goal of floor plan generation is to automatically orchestrate the elements intrinsic to the floor plan (\eg rooms, doors, walls) into a reasonable composition. 
Rule-based~\cite{muller_procedural_2006, peng_computing_2014, yin_generating_2022} and learning-based~\cite{wu_data-driven_2019, nauata_house-gan_2020, shabani_housediffusion_2023, tang_graph_2024} methods exists to do so. 
We categorize the literature into three distinctive approaches. First, boundary-constrained floor plan generation methods~\cite{peng_computing_2014, wu_data-driven_2019, sun_wallplan_2022}, constrain the generative process on the external walls that separate the interior of the building from the outside. 
Second, graph-constrained floor plan generation methods~\cite{nauata_house-gan_2020, nauata_house-gan_2021, luo_floorplangan_2022, yin_generating_2022, tang_graph_2024} allow for fine-grained control of the floor plan by constraining the generative process on the functional diagram, which is usually represented as a graph. 
Instigated by~\cite{nauata_house-gan_2020}, especially graph-constrained floor plan generation of residential houses has led to a broad range of domain-specific network architectures and optimization frameworks:
Conv-MPN~\cite{zhang_conv-mpn_2020} was introduced to better capture topological and shape-wise features simultaneously; generative adversarial networks (GANs) over graphs to enable graph-structured generation~\cite{nauata_house-gan_2020, nauata_house-gan_2021}; discrete diffusion models~\cite{shabani_housediffusion_2023} to accommodate the denoising of geometrical shapes; transformer GANs~\cite{tang_graph_2024} to capture both local and global relations across nodes in the graph; etc. 
Third, boundary- and graph-constrained floor plan generation methods~\cite{hu_graph2plan_2020, upadhyay_flnet_2022} allow control over the boundary as well as the graph, a setting that is closest to most real-world design conditions. 
Besides the graph and boundary, we constrain the generative process on the other necessary structural components (\eg, load-bearing walls).

\section{Dataset of Modified Swiss Dwellings (MSD) }\label{sec:data}

MSD is derived from the Swiss Dwellings dataset (SD)~\cite{standfest_swiss_2022}.
We carefully cleaned and curated SD by taking the following steps:

\begin{itemize}
    \item[$\bullet$] \textbf{Feature removal.} 
    All non-floorplan geometries are removed (\eg "bathtub", "stairs"; see the full list in the suppl. mat.).
    
    \item[$\bullet$] \textbf{Residential-only filtering.} 
    We remove floor plans that include non-residential-like geometric details (\eg areas categorized as "waiting room", "dedicated medical room"; see the full list in the suppl. mat.). 
    This led to the removal of 2,305 (16.6\%) floor plans.
    
    \item[$\bullet$] \textbf{Near-duplicate removal.} 
    Many floor plans that come from the same building stem from the same plan ID~\cite{standfest_swiss_2022} (see suppl. mat. for details on ID nesting in SD). 
    Floor plans with the same plan ID are based on the same layout, indicating that the spatial arrangements are nearly identical. 
    Hence, we sample only one-floor plan per plan ID to drastically reduce the amount of near-duplicates. 
    Specifically, we sample the floor plan with the lowest elevation. 
    This led to the removal of 4,395 (31.6\%) floor plans.
    
    \item[$\bullet$] \textbf{Medium- to large-scale filtering.} 
    Floor plans are removed that contain fewer than 15 areas. In addition, every floor plan should have at least two "Zone 2"-categorized areas. 
    This led to the removal of 1,541 (11.1\%) floor plans.
\end{itemize}

\noindent 
Additional steps for cleaning and filtering are provided in the suppl. mat., leading to the removal of an extra 388 (2.8\%) floor plans. Ultimately, the amount of floor plans in MSD is 5,372.

\begin{figure*}[t]
    \centering
    \includegraphics[width=\textwidth]{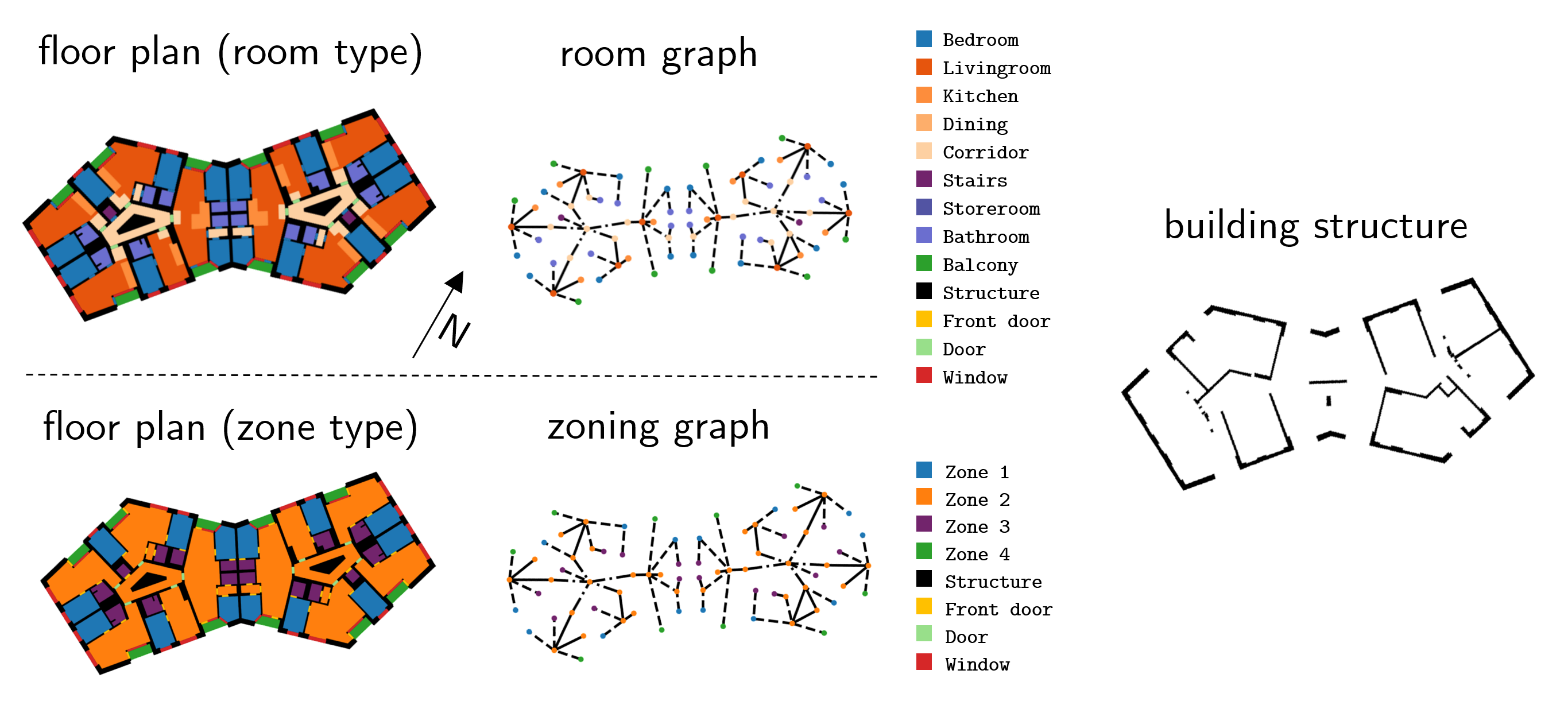}
    \small
    \caption{
        \textbf{Representation of floor plan data.} 
        MSD contains three types of data.
        (\textbf{Left}) The floor plan is the complete layout of the floor of the building, including the position and shape of the rooms, doors, and windows.
        The areas are labeled by color (see legends). 
        There are two category systems: 1) category is based on the type of the area ('room type'), in which each area has a room type (\eg, "bedroom"); and 2) category is based on the zone of the area ('zone type'), in which each areas has a zone type (\eg, "zone1").
        (\textbf{Middle}) The associated room and zoning graphs are depicted on the right of the floor plans. 
        The node colors are equivalent to the colors of the floor plan. 
        The position of each node is taken as the centroid of the area that the node represents.
        (\textbf{Right}) A binary image of the necessary structural components of the floor plan.
        }
    \label{fig:data}
\end{figure*}


\paragraph{\textbf{Categorization \& labeling}.} 
We refer to an area as any well-defined part in a floor plan that a person could walk in or through (\eg, bedroom, corridor, balcony.)
Each area has three attributes: 1) the shape of the room (represented as polygon), 2) a room type category, and 3) a zoning type category. 
The zoning types (or zones) are based on the categorization made in~\cite{khodabakhshi_procedural_2022}: "zone1" refers to a private space, "zone2" to a public space, "zone3" to a service space, and "zone4" to an outside place.

\paragraph{\textbf{Image extraction}.}
The floor plan images are made by 'drawing' the floor plan's corresponding geometries on a single-channel image canvas. 
The room category labels are represented as integers (\eg 0 for "living room"). 
The coordinates $x$ (east-to-west) and $y$ (south-to-north) are defined in meters. 
The floor plans are mostly centered around $(x, y){=}(0, 0)$. 
To retain the information of the original coordinates within the image representation, two extra channels are added on top of the image (see suppl. mat. for details).

\paragraph{\textbf{Graph extraction}.}
An (access) graph is an attribute of a floor plan and explicitly models the connections (edges) between the areas (nodes). We use an algorithmic approach to extract the graphs from the room shapes. The procedure is as follows:

\begin{itemize}
    \item[$\bullet$] 
    \textbf{Edge types.} We define three types of edge connectivity: 1) "passage" when one can walk from one area to the other without a door in between; 2) "door" if two areas are connected by a door; and 3) "front door" if two areas are connected by a front door.
    
    \item[$\bullet$] 
    \textbf{Edge development.} We iterate over all possible area pairs and create an edge between the two areas if 1) either the polygons that define the shapes of the areas are close enough ($\leq$ 0.04 m) -- in this case the edge type is "passage" -- or 2) there is a door for which the polygon that defines the door's shape is close enough to both area shapes ($\leq$ 0.05 m) -- in this case the edge type is either "door" or "front door" depending on the type of door.
    
    \item[$\bullet$] 
    \textbf{Node development.} We include all necessary geometric and semantic information as node attributes. "centroid": is the center of the area. "geometry": is an array of the 2D coordinates representing the shape of the room as a polygon. "roomtype": is an integer representing the room category of the area. "zonetype": an integer representing the zoning category of the area. 
    
    \item[$\bullet$] 
    \textbf{Room and zoning graph.} We define the room graph as the graph including only "roomtype" node attributes, and the zoning graph as the graph including only "zonetype" node attributes. Fig.~\ref{fig:data} depicts both room and zoning graphs. 
\end{itemize}

\paragraph{\textbf{Structure extraction}.}
Structural elements of the building are often regarded as fixed beforehand. 
The structural elements in floor plans include load-bearing walls and columns. 
Accordingly, the structural elements of the floor plans are extracted and regarded as a part of the input to frame a plausible design problem. 
The criteria to distinguish the structural walls from the regular separating walls are based on the base wall thickness. 
The base wall thickness for each floor plan is the 60\% quantile of the full set of existing walls’ thickness given a floor plan. 
Any wall with a thickness larger than the base thickness value is then regarded as a load-bearing wall. 
Similar to load-bearing walls, columns are regarded as fixed as well. 
Hence, all geometric details categorized as "columns" are appended to the building structure.

\subsection{Comparison to other datasets}

We compare MSD to RPLAN~\cite{wu_data-driven_2019} and LIFULL~\cite{lifull_co_ltd_lifull_nodate}. \tabl{stats} accompanies the findings that are given next.

\paragraph{\textbf{Origin}.} 
RPLAN and LIFULL contain floor plans that originate from, resp., the Asian and Japanese houses. 
MSD, on the other hand, is the first large-scale and detailed floor plan dataset originating from Europe, specifically Switzerland. 
While investigating the differences between Asian and European floor layouts is in itself an interesting endeavor (and goes beyond this paper), with region-specific floor plan datasets, machine learning algorithms can assist in designing buildings that cater to specific cultural preferences or comply with local building codes.
We are actively extending the current dataset, to dwellings from other regions in Europe as well.

\paragraph{\textbf{Realism}.} 
The vectorized floor layouts in LIFULL are extracted from the original image dataset~\cite{lifull_co_ltd_lifull_nodate} using the vectorization algorithm proposed in~\cite{liu_raster--vector_2017}. 
This vectorization algorithm is not necessarily error-proof and reaches on average an accuracy ($1 - \text{IoU}$) of 88.5 and 94.7 for predicting the room shapes and wall junctions, resp. (Tab.~1 in~\cite{liu_raster--vector_2017}). 
Furthermore, floor plans in RPLAN and LIFULL are re-oriented to make them axis-aligned, arguably for easy processing and use. 
MSD, on the contrary, retains the orientation of the floor plans -- a feature of significant importance to the quality of the floor plan design.

\paragraph{\textbf{Complexity}.} 
RPLAN and LIFULL comprise floor plans of isolated apartments; therefore, the data do not contain information about the relations between the distinct apartments. 
MSD is the first large-scale floor plan dataset of multi-apartment dwellings; hence, the connections between the distinct apartments are explicitly modeled (\tabl{stats}, column 5). 
Moreover, rooms in RPLAN and LIFULL are Manhattan-shaped. 
In the real world, rooms are often more diverse in shape (\ie, non-Manhattan (NM)). 
MSD retains the imposed shapes of the rooms, even if the shapes are NM (\tabl{stats}, column 6). 
Additionally, rooms in MSD consist of more corners (\tabl{stats}, column 3).

\begin{table}[ht]

    \caption{
    \small
    \textbf{MSD compared to RPLAN~\cite{wu_data-driven_2019} and LIFULL~\cite{lifull_co_ltd_lifull_nodate}}.
    \textbf{Complexity} is measured by the average number of corners per room (c$_3$), rooms per unit (c$_4$), and units per floor (c$_5$). Additionally, c$_6$ indicates a \textit{significant} share ($>5\%$) of NM-shaped rooms.
    \textbf{Information} is measured per the existence of room labels (room type and zoning type in c$_7$ and c$_8$, resp.) and of doors (c$_9$).
    \textbf{Size}: c$_{10}$ and c$_{11}$ provide the \textit{total} number of rooms and units in the datasets, resp. Note that the \textit{total} number reflects the reduced dataset sizes when near-duplicates are removed.
    \textbf{Diversity} is measured as the entropy, $\mathrm{H}_g$ (see~\eq{graph-entropy}), over the distribution of graphs (c$_{12}$). MSD sets a new standard as a more complex and realistic floor plan dataset. Note that c$_i$ is column $i$.
    }
    
    \centering
    \resizebox{1\linewidth}{!}{%
    \begin{tabular}{l c cccc ccc cc c}
        \toprule
        
        \multirow{2}{*}{\textbf{Dataset}} & 
        \multirow{2}{*}{\textbf{Origin}} &
        \multicolumn{4}{c}{\textbf{Complexity}} & 
        \multicolumn{3}{c}{\textbf{Information}} &
        \multicolumn{2}{c}{\textbf{Size}} &
        \textbf{Diversity}
        \\
        
        & & $\frac{\text{corners}}{\text{room}}$ & $\frac{\text{rooms}}{\text{unit}}$ & $\frac{\text{units}}{\text{floor}}$ & NM 
        & function & zoning & doors 
        & \# rooms & \# units 
        & $\mathrm{H}_g$ 
        \\
        
        \midrule

        LIFULL  
        & Asia 
        & 4.54 & 8.15 & 1.00 & \xmark 
        & \cmark & \xmark & \xmark 
        & \underline{\textbf{489.4.3K}} & \underline{\textbf{61.3K}}
        & 7.79
        \\
        
        RPLAN  
        & Asia 
        & 5.04 & 6.67 & 1.00 & \xmark 
        & \cmark & \xmark & \cmark 
        & 161.8K & 24.2 K
        & 4.56
        \\
        
        \midrule
        
        MSD (ours) 
        & Europe
        & \underline{\textbf{8.68}} & \underline{\textbf{8.75}} & \underline{\textbf{3.52}} & \cmark 
        & \cmark & \cmark & \cmark 
        & 163.5K & 18.5K 
        & \underline{\textbf{8.02}}
        \\
        \bottomrule
    \end{tabular}

    \label{table:stats}
        
    }
\end{table}

\paragraph{\textbf{Information}.} 
Where RPLAN and LIFULL comprise floor plans in either image (RPLAN) or vectorized (LIFULL) format, MSD explicitly represents the floor plans in an image, vectorized, \textit{and} graph formats. 
On top of the full floor plan layouts, MSD contains the corresponding structural necessary components, represented as binary images. 
In addition to room type labels, MSD provides the zoning category of each room as well.
Floor plans in LIFULL do not contain door information. 
MSD contains (as does RPLAN) the geometric information of the doors as well -- a necessary feature to better understand and exploit the interconnectivity between spaces.

\paragraph{\textbf{Size \& diversity}.}
RPLAN ($\sim$81K) and LIFULL ($\sim$124K) have significantly more floor plans than MSD ($\sim$5K) (column 8 in~\tabl{stats}). 
The total number of \textit{rooms} is much closer though: $\sim$165K for MSD vs. $\sim$539K and $\sim$1010K for RPLAN and LIFULL, resp.
However, RPLAN and LIFULL contain a serious amount of near-duplicates.
We measure the number of near-duplicates in a floor plan dataset by measuring the $\text{MIoU}$ between pairs of floor plans and removing those that exceed a certain threshold, which we set to 0.87, equivalent to the procedure in~\cite{van_engelenburg_ssig_2023}. 
While the number of near-duplicates in MSD is negligible (<1\%), those in RPLAN and LIFULL are not: 70\% for RPLAN and 50\% for LIFULL. 
When filtering out the near-duplicates from the original RPLAN and LIFULL datasets, the number of rooms in MSD and RPLAN are approximately equal, while the number in LIFULL still remains significantly larger. 
In terms of the topology of the room graphs, MSD is notably more diverse than RPLAN and slightly more diverse than LIFULL. 
The diversity is measured as the entropy over the distribution of graphs: 

\begin{equation}\label{eq:graph-entropy}
    \mathrm{H}_g = - \sum_{g \in \mathcal{G}} p_g(g) \log p_g(g)
\end{equation}

\noindent where $\mathcal{G}$ is the set of distinct un-attributed room graphs in the dataset and  $p_g(g)$ is the probability of a floor plan having a corresponding room graph equal to graph $g$. 
$p_g(g)$ is numerically approximated through the dataset.
The entropy over the graph distributions for MSD, RPLAN, and LIFULL are resp., 8.02, 4.56, and 7.79, which reveals that MSD is most diverse in terms of the connectivity.

In summary, MSD is a large floor plan dataset of European (Switzerland) building complexes and exceeds other datasets in layout complexity and graph diversity. 
MSD is, above all, the first big dataset that makes explicit the inter-relations between spatially-connected apartments. 
\fig{motivational} further reveals the noteworthy differences between MSD and other floor plan datasets.
\section{Floor plan generation task}\label{sec:task}

We set our task as a real-world design formulation by bridging the schematic design (spatial zoning) to the detailed design (floor layout). 
Similar to~\cite{hu_graph2plan_2020}, we formulate the floor plan generation task as a multi-modal constrained optimization problem with the following in- and outputs:

\begin{itemize}
    \item[$\bullet$] \textbf{Input 1.} 
    The building structure indicates where the necessary structural components are positioned. 
    The building structure is represented as a set of geometries or as a binary image.

    \item[$\bullet$] \textbf{Input 2.} 
    The zoning graph defines the connectivity of areas and is represented as a graph with category-attributed nodes and edges that indicate the zoning classes.

    \item[$\bullet$] \textbf{Output.} 
    The floor plan which is either represented as an image with pixel values that indicate the room classes or a room graph with geometry- and category-attributed nodes that indicate the shape and room category. 
    We include both representations to enable the use of different model architectures such as convolutional neural networks (that need images) and graph neural networks (that need graphs).
\end{itemize}

\subsection{Evaluation metrics} 

To measure the performance of the models, we compute both the visual and topological similarities between the predicted and ground truth floor plans. The ordered sets of target and predicted floor plans are, resp., denoted as $\mathcal{Q} = \left\{Q_i\right\}_{1, \ldots, N}$ and $\mathcal{K}= \left\{K_i\right\}_{1, \ldots, N}$, in which $N$ is the size of the test set.

\paragraph{\textbf{Mean Intersection-over-Union}.} 
The rooms of a floor plan must have the correct shape and location.
To measure the performance at the pixel level, the Mean Intersection-over-Union (MIoU) between $\mathcal{Q}$ and $\mathcal{K}$ is used. Equivalent to \cite{van_engelenburg_ssig_2023}, MIoU across all classes $c \in \mathcal{C}$ is computed by:

\begin{equation}\label{eq:miou}
    \mathrm {MIoU}(\mathcal {Q};\mathcal {K})= 
    \frac{1}{N} 
    \sum _{i=1}^{N} 
    \sum _{c \in \mathcal{C}} 
    \frac{R_c(Q_i) \bigcap R_c(K_i)}{R_c(Q_i) \bigcup R_c(K_i)}, 
\end{equation}

\noindent where $R_c(\cdot)$ is the function that outputs the region in the image occupied by $c$.

\paragraph{\textbf{Graph compatibility}.} 
It is similarly important that the topology of the predicted floor plan's composition closely matches that of the ground truth. 
To measure the consistency between the predicted and target graph, we compute the graph compatibility between the graph extracted from the predicted floor plan and the target graph. 
\cite{nauata_house-gan_2020, nauata_house-gan_2021, shabani_housediffusion_2023, tang_graph_2024, van_engelenburg_ssig_2023} measure the compatibility based on a graph edit distance (GED)~\cite{sanfeliu_distance_1983}. 
The output of the floor plan generation task, however, hinders the practical use of the GED in our case, because doors are not predicted in our setting. 
Similar to the graph extraction algorithm used in the making of MSD, we extract the room graphs of the predicted floor plan by looping through all the pairs of different areas of a given floor plan. 
We assign an edge whenever the minimum distance between the areas is less or equal to a buffer. 
The compatibility is computed by checking whether the edges from the target graph are retained in the predicted graph:

\begin{equation}\label{eq:compatibility}
    \mathrm{Compatibility} (\mathcal {Q};\mathcal {K}) = 
    \frac{1}{N} 
    \sum _{i=1}^{N} 
    \frac{1}{\left|\mathcal{E}^{k}_{i}\right|} 
    \sum _{e \in \mathcal{E}^{k}_{i}} 
    \mathbf{1} \left[ e \in \mathcal{E}^{q}_{i} \right], 
\end{equation}

Extracting graphs from noisy pixel maps is error-prone; hence, we refrain from using it for methods alike (\eg, for UN; see~\sect{unet}).
For graph-based methods (\eg, for MHD; see~\sect{mhd}), graph extraction from the predicted layouts could lead to errors as well, usually when a predicted layout contains many overlapping rooms.
However, we found that such scenarios do not often occur (see suppl. mat. for details).
Hence, we deem extracting room edges algorithmically as reasonable.
(Note that previous works use a similar algorithmic approach too.)

As mentioned before, previous works use a GED to measure the compatibility~\cite{nauata_house-gan_2020, nauata_house-gan_2021, shabani_housediffusion_2023, tang_graph_2024, van_engelenburg_ssig_2023}. 
Hence, lower scores suggest better methods. We measure a graph \textit{similarity} instead of \textit{distance}. 
Therefore, a high score (instead of low in the case of GED) positively correlates with performance.
\section{Models}\label{sec:experiment}
We develop two baseline models to benchmark MSD: a diffusion- and segmentation-based approach.
\fig{methods} provide visual clarifications of both baselines. 
We also tested the generalizability of \textit{HouseGAN++}~\cite{nauata_house-gan_2021} and \textit{FLNet}~\cite{upadhyay_flnet_2022}, for which the results are provided in the suppl. mat.
(Note that most of the details on the model architectures, the training, and pre-processing are given in the suppl. mat.)

\subsection{Modified HouseDiffusion (MHD)}\label{sec:mhd}
\begin{figure*}[t]
    \centering
    \includegraphics[width=\textwidth]{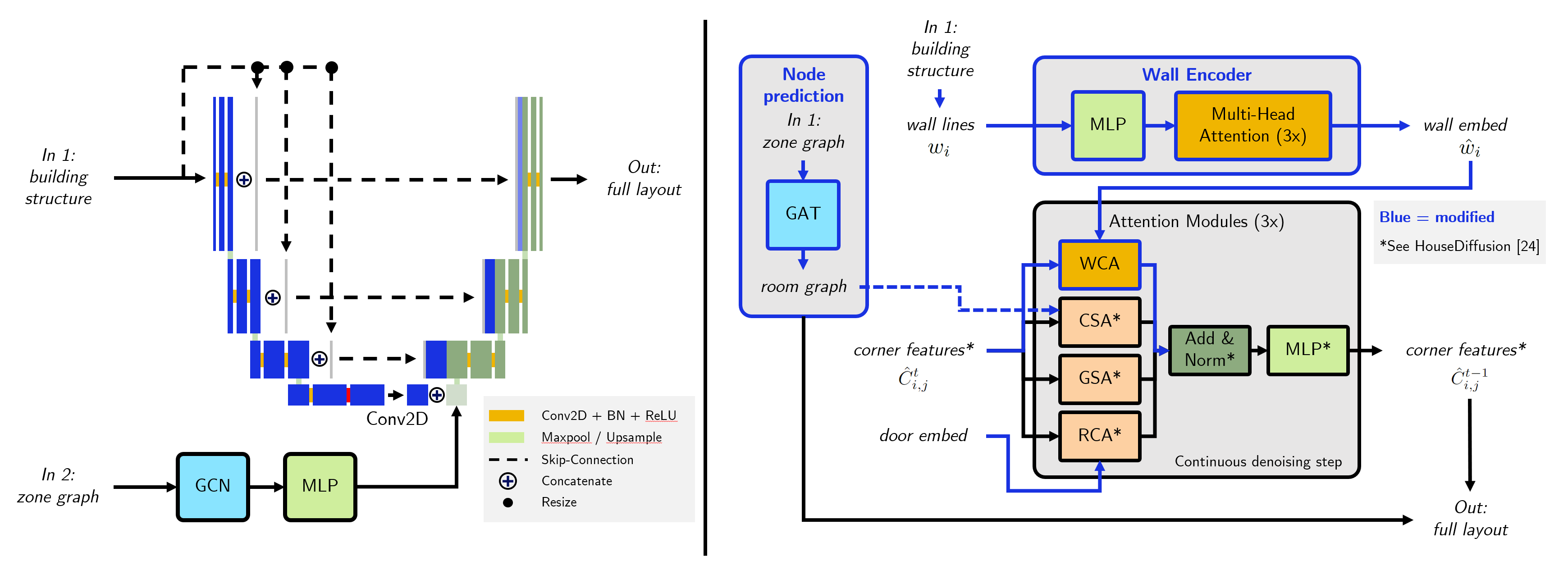}
    \small
    \caption{
        \textbf{Baseline methods for floor plan generation.} 
        (\textbf{Left: UN}) UN takes the building structure (image) as input to the U-Net. 
        The U-Net is composed of an encoder and decoder using the conventional up- and down-sampling 2D convolutions, resp., and includes skip connections between the encoder and decoder feature maps at equivalent feature map scales. 
        A GCN is used to map the zoning graph to a feature vector which is concatenated to the latent space of the U-Net. 
        (\textbf{Right: MHD}) A wall encoder is used to map the pre-processed building structure into corresponding wall embeddings. 
        MHD expands HD~\cite{shabani_housediffusion_2023} by introducing an extra attention module (WCA) between the wall embeddings from and corner features of the rooms. 
        A GAT is separately trained to predict the room types from the zoning types, which are used to "color" the full layout. 
        }
    \label{fig:methods}
\end{figure*}

\textit{HouseDiffusion} (HD)~\cite{shabani_housediffusion_2023} is a state-of-the-art model for graph-constrained floor plan generation. 
To adapt HD to our task, a cross-attention module is added, which effectively conditions the diffusion process on the building structure. 
To learn the room graph from the zoning graph, we use a GAT~\cite{velickovic_graph_2018}, which operates as a pre-processing step to the diffusion process. We coin our method \textit{Modified HouseDiffusion} (MHD).
\fig{methods} (right) provides all modifications and added modules, which are depicted in blue.

\paragraph{\textbf{HouseDiffusion (HD)}.}
In HD, floor plans are represented by a set of polygons $P = \{P_1, P_2, ..., P_N\}$, each representing a room or door. 
Each polygon $P_\bullet$ is defined by a sequence of 2D corner coordinates, $C_{l, m} \in \mathbb{R}^2$ in which $l$ refers to $l$-th polygon and $m$ the index of the corner. 
In the forward process, noise is added to the corner coordinates at each timestep $t$ such that at timestep $t{=}T$ all corner coordinates follow a normal distribution. 
The corner coordinates at timestep $t{=}0$ remain unaltered. 
The goal of the model is then to learn the reverse process (\ie, to iteratively denoise the noisy corner candidates back). 
At its core, HD consists of three attention layers with structured masking: 
1) CSA, limiting attention among corners in the same room or door, 
2) GSA, full stack attention between every pair of corners across all rooms, 
and 3) RCA, limiting attention between connected room-to-door corner pairs.

\paragraph{\textbf{Wall-cross attention (WCA)}.}
MHD expects the building structure to be encoded as a set of wall elements (straight lines) $w_i$, which are extracted from the binary image by a morphological thinning technique followed by skeletonization (see suppl. mat. for details).
Each wall element is converted into a 512-d wall embedding $\hat{w}_i$ by an MLP followed by three multi-head attention modules. 
To condition the model on the building structure, we add an extra cross-attention module (WCA) between all corner and wall embeddings. 

\paragraph{\textbf{Graph attention network (GAT)}.}
Instead of changing HD's architecture to denoise a room type for each corner in addition to the coordinates, we separately learned the room types beforehand. 
We use a GAT~\cite{velickovic_graph_2018} to learn the room graph from the zoning graph, by essentially framing the problem as node classification.

\paragraph{\textbf{Minimum rotated rectangle (MRR)}.}
In HD, the number of corners is sampled from the known corner count distributions per room type in the training set. 
In contrast to RPLAN, which typically has between 4 and 10 corners per room, MSD contains many areas with a much larger amount of corners, making it more difficult for the model to appropriately denoise the polygons. 
In addition to doing experiments with the full set of corners (POL) we approximate the polygons by a minimum rotated rectangle fit (MRR), and subsequently learn to denoise the MRRs instead. 

\subsection{Graph-informed U-Net (UN)}\label{sec:unet}
We propose a floor plan generation model based on U-Net~\cite{ronneberger_u-net_2015} for direct prediction at the pixel level. 
At the deepest level of the network, the U-Net is constrained on a graph-level encoding of the zoning graph which is learned by a GCN~\cite{kipf_semi-supervised_2017}.

\paragraph{\textbf{U-Net}.} 
A U-Net is used to 'segment' the building structure into the floor plan. 
Essentially, a U-Net is an autoencoder with the addition of skip-connection between the feature maps of the encoder and decoder that share the same feature map resolution. 
Similar to the original U-Net implementation, we use convolutional layers for both down- and up-sampling. The output of the U-Net is the floor plan image.

\paragraph{\textbf{Graph convolutional network (GCN)}.}
A GCN~\cite{kipf_semi-supervised_2017} is used to learn a fixed-sized graph-level embedding of the zoning graph. 
To effectively combine the graph and boundary representations, the graph-level embedding is concatenated to the deepest layer's feature map of the U-Net. 

\paragraph{\textbf{Boundary pre-processing}.}
Inspired by~\cite{wu_data-driven_2019}, we pre-process the building structure into a 3-channel image, distinguishing the interior (channel 1), the exterior (channel 2), and the original building structure (channel 3). 
We use Segment Anything~\cite{kirillov_segment_2023} to extract the interior and exterior from the building structure. 
\section{Results}

\paragraph{\textbf{MIoU}.}
To measure the visual similarity, we use (Eq.~\ref{eq:miou}).
The polygonal outputs of MHD are drawn on the same image canvas as that of the ground truth. 
The drawing order matters and is from largest to smallest region size to make sure the smaller areas are not entirely occluded by larger areas. 

\begin{figure*}[t]
    \centering
    \includegraphics[width=\textwidth]{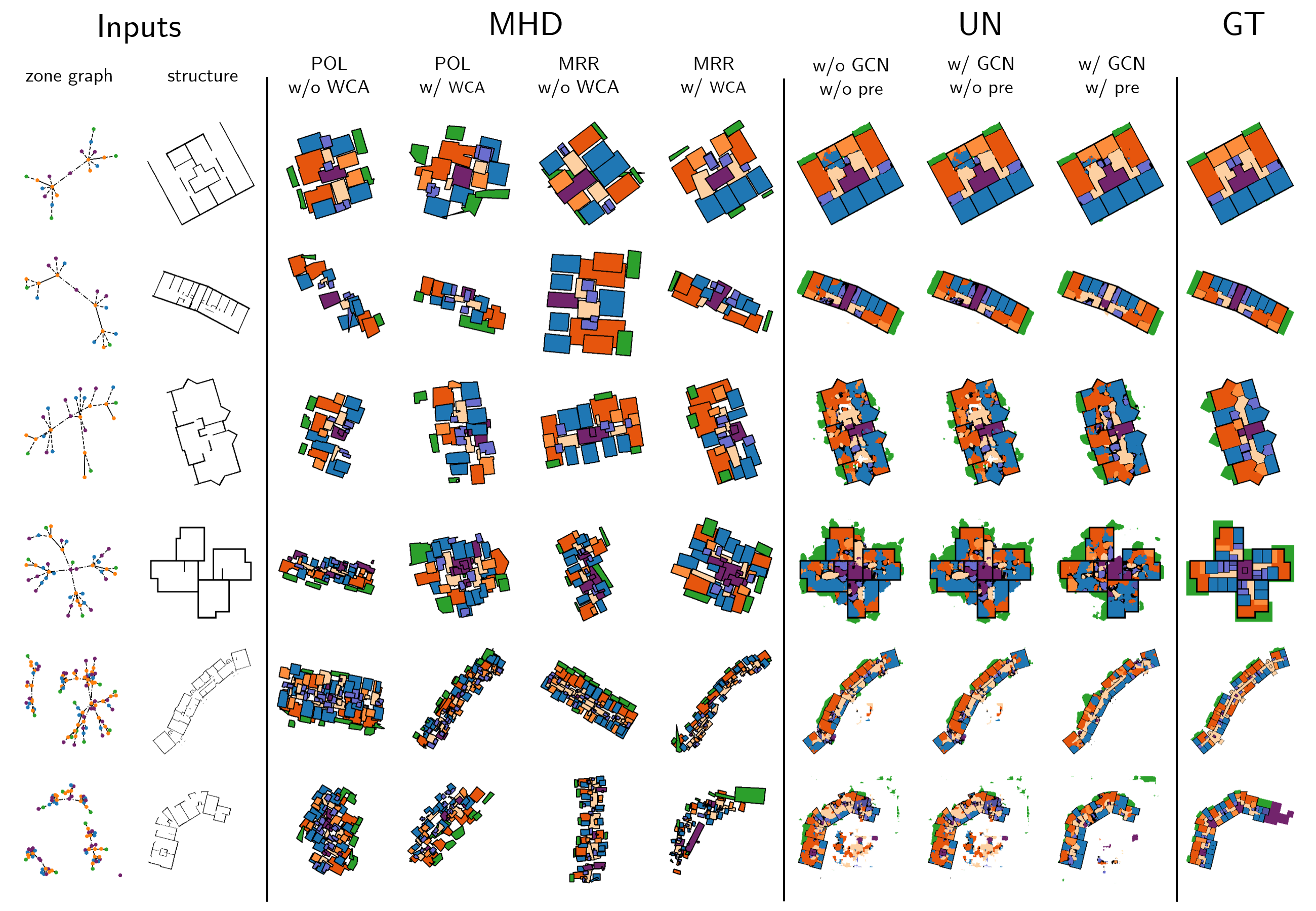}
    \small
    \caption{
        \textbf{Example generations of MHD and UN.}         
        Columns 1 and 2 show the inputs: the zone graph and building structure respectively. Columns 3 - 6 show the floor plans generated by the MHD variants. Columns 7 - 9 show the floor plans generated by the UN variants. Column 10 shows the ground truth.
        }
    \label{fig:experiments}
\end{figure*}

The average MIoU (\tabl{experiments}) ranges from 10.9 to 42.4. 
Intuitively, this means that for all the models, a pixel is more likely to be predicted incorrectly.
We conclude that the performance is poor and does not yet comply with the performance standards we would like to see. 
The apparent mismatch between predictions and targets is further investigated based on some of the generated examples, provided in~\fig{experiments}. 
Even though the relative positioning of the areas (for MHD models) and pixel regions (for UN models), for most generated floor plans, are predicted quite accurately, the exact locations and precise shapes of the areas and regions are far from accurate yet -- explaining, indeed, the low overlap scores. 

\begin{table*}[ht]
    \small
    \centering
    \caption{
        \textbf{MIoU and compatibility scores for MHD and UN.}. 
        The best scores across indicated in bold, and we underline the scores that are best within each approach. 
        The scores for different floor plan 'sizes' -- based on the range in a number of areas -- are provided in the different columns of MIoU and graph compatibility. 
        The graph compatibility scores for the UN-based models are not available (n.a.) because graphs cannot be reasonably extracted from the output images. 
        The vanilla version of UN is denoted as "U-Net", and "(pre)" indicates the use of Segment Anything for pre-processing. MHD considers either the full polygons (POL) or a minimum rotated rectangle fit (MRR). 
        "+WCA" indicates the use of the full-stack attention module between corner and wall embeddings.
    }
    \resizebox{\textwidth}{!}{
    \begin{tabular}{l|cccccc|cccccc}
        \toprule
        
        \multirow{2}{*}{\textbf{Method}}          & \multicolumn{6}{c|}{\textbf{MIoU} ($\uparrow$)}                     & \multicolumn{6}{c}{\textbf{Compatibility} ($\uparrow$)} \\
        
                                         & avg. & 15 -- 19 & 20 -- 29 & 30 -- 39 & 40 -- 49 & 50+     & avg. & 15 -- 19 & 20 -- 29 & 30 -- 39 & 40 -- 49 & 50+  \\
        
        \midrule
        
        U-Net                            & 32.5 & 33.4     & 33.1     & 32.8     & 29.7     & 29.3    & n.a. & n.a.     & n.a.     & n.a.     & n.a.     & n.a. \\
        UN                     & 40.6 & 44.8     & 42.9     & 38.4     & 32.3     & 30.4    & n.a. & n.a.     & n.a.     & n.a.     & n.a.     & n.a. \\
        UN (pre)            & \textbf{\underline{42.4}} & \textbf{\underline{45.4}} & \textbf{\underline{45.4}} & \textbf{\underline{40.6}} & \textbf{\underline{35.1}} & \textbf{\underline{32.2}}  & n.a. & n.a.  & n.a.  & n.a.  & n.a.   & n.a. \\
        
        \midrule 
        
        MHD (POL)             & 10.9 & 11.6	& 11.5 &	10.2 &	9.8 &	9.1 &   80.3 & 80.1	& 79.5	& 81.4 &	80.4	& 81.9\\
        MHD (POL) + WCA      & 17.9 & 18.6 &	18.4	& 17.6	& 16.2	& 15.5 & 71.1 & 70.5	& 70.7	& 71.4	& 71.9	& 73.7  \\
        MHD (MRR)             & 11.5 & 12.2	& 12.2	& 11.1	& 10.2	& 9.0 &   \textbf{\underline{87.1}} & \textbf{\underline{85.9}} & \textbf{\underline{87.3}}     & \textbf{\underline{88.0}}     & \textbf{\underline{87.5}}     & \textbf{\underline{88.6}} \\
        MHD (MRR) + WCA      & \underline{21.8} & \underline{23.5}     & \underline{22.0}     & \underline{21.0}     & \underline{20.1}     & \underline{17.9}    & 76.2 & 76.0     & 75.6     & 75.8     & 77.6     & 79.2 \\
        \bottomrule
    \end{tabular}

    \label{table:experiments}
    
    }
\end{table*}

\paragraph{\textbf{Graph compatibility}.}
The compatibility is only measured for MHD, because reliably extracting graphs from the floor plans generated by UN is too ambiguous. 
For an image size equal to 512, the buffer is set to 5 to allow some, but not too much, space between rooms. 
The compatibility is computed between the extracted room graph of the predicted floor plan and the room graph of the target floor plan by Eq.~\ref{eq:compatibility}. 

Compared to the MIoU, the compatibility scores, ranging between 74.4 and 87.0, are much higher; thus, the topology of the zoning graphs is largely retained in the generated floor plans. 
Therefore, MHD can reasonably well learn how the rooms should be composed.

\paragraph{\textbf{UN vs. MHD}.}
\fig{experiments} shows the qualitative differences between MHD and UN models. 
One observation is that MHD creates composed shapes in which the distinct areas can be easily separated by the eye. 
In contrast, UN models create segmented scenes for which it is less visible which set of pixels belongs to which room. 
The MIoU scores are, however, much higher for the UN models from which we can conclude that the UN models have a better understanding of the placement of specific rooms in relation to the building structure. 
Indeed, the example outputs of the UN models clearly show that the central regions in the floor plans are usually corridors, that the balconies are placed outside the building structure, and that the kitchen is usually located close to the living room -- characteristics that are to lesser extend present in the floor plans generated by MHD.

The discrepancy in the performance (in MIoU) might stem from the different losses. 
Clearly, the loss of UN (cross-entropy at pixel level) is closely aligned with evaluating MIoU. 
However, in the case of MHD, the loss and evaluation are not necessarily as closely aligned.
The objective is similar to other diffusion models: at each iteration, you randomly select a time-step $t$ and learn a mapping for the reverse noise, which is parameterized by a neural network as $e_{\theta} \left( C_{l,m}, t\right)$.
Hence, the neural network $e_{\theta} \left( \bullet, \bullet \right)$ learns to effectively denoise corner points \textit{for a given time step}.
This is not necessarily the same as learning a mapping from input (structure and graph) to a \textit{fixed} output (floor plan layout), which could for a part explain the discrepancy in performance.

\subsection{Ablation studies}

\paragraph{\textbf{UN}.}
From~\tabl{experiments}, the effects of adding the GCN and/or pre-processing are significant and increase MIoU. 
The impact of the GCN is most significant for smaller building complexes. 
The MIoU scores for larger floor plans are comparable across the three methods, which suggests that the GCN struggles with larger graphs, emphasizing the need for graph models that can cope better with both small- as well as large graphs. 
The examples in \fig{experiments} show that the addition of the pre-processing tends to improve the placement of the areas within the interior of the building.  

\paragraph{\textbf{MHD}.}
Observed from the generated examples in \fig{experiments}, adding WCA leads to floor plans that follow the building structure reasonably well. 
This is also shown by the increase in MIoU scores between the models with and without WCA. 
However, the addition of WCA leads to degraded graph compatibility, likely because when conditioning on the building structure, the model has to learn to place rooms along the existing structure, instead of only placing rooms relative to each other. 
The model that uses the minimum rotated rectangle (MRR) approximation performs better than the model with full polygons (POL), both on MIoU and compatibility, which we attribute to the following potential causes. 
First, some rooms in MSD have many corners and are likely more complicated to learn. 
Second, the number of corners for POL is sampled independently of the building structure, which can lead to room polygons having too few or too many corners to fit the building structure. 

Overall, the floor plans often look infeasible. 
We could, however, train MHD on RPLAN successfully (see suppl. materials); hence, we believe that the poor results do \textit{not} come from improper training.
Instead, we attribute the poor results to the more complex benchmark we set: more complex graphs; more irregularly shaped rooms; unit connectivity; no axis alignment; etc.

\section{Conclusion}
We developed MSD -- a large-scale floor plan dataset of building complexes.
In contrast to the other floor plan datasets, MSD contains more complex floor plans. 
To test the generalizability and scaling of the current state-of-the-art floor plan generation method to MSD, we developed two baseline models.
The baseline models were highly inspired by previous works and only altered where needed.  
Our experiments show that the generation of more complex, hence more realistic, floor plans cannot yet be properly addressed by strategies that are currently most promising in floor plan generation.
To address real-world floor plan design, our benchmark asks for even smarter methods in the future.

\smallskip

\noindent \textbf{Acknowledgment:}
We would like to thank all architects and students who participated in our user study (see suppl. mat.).

\bibliographystyle{splncs04}
\bibliography{main}

\clearpage
\setcounter{page}{1}

\chapter*{Supplementary Materials: MSD: A Benchmark Dataset for Floor Plan
Generation of Building Complexes}

\section{Dataset of MSD}

\subsection{Categorization and ID nesting} \label{subsec:nesting}

The Swiss Dwellings dataset (SD)~\cite{standfest_swiss_2022} is stored as a large dataframe\footnote{A dataframe is defined as a two-dimensional data structure, for which the naming is borrowed from the \href{https://pandas.pydata.org/}{Pandas library} in Python.}. 
Each row in the dataframe corresponds to a geometrical detail \eg~living room, wall element, or balcony. 
The entity has a \textit{type} and, further nested, \textit{subtype} categorization (which form two of the columns in the dataframe):

\begin{itemize}
    \item The \textbf{"feature"} type includes the following subtypes: "washing machine", "shower", "bathtub", \textcolor{blue}{"kitchen"}, \textcolor{blue}{"elevator"}, "built in furniture", "stairs", "toilet", "sink", "ramp".
    \item The \textbf{"separator"} type includes the following subtypes: "wall", "railing", "column".
    \item The \textbf{"opening"} type includes the following subtypes: "entrance door", "window", "door".
    \item The \textbf{"area"} type includes the following subtypes:: \textcolor{red}{"radiation therapy"}, \textcolor{red}{"office"}, "corridors and halls", "wintergarten", \textcolor{red}{"salesroom"}, \textcolor{red}{"studio"}, \textcolor{red}{"open plan office"}, \textcolor{red}{"outdoor void"}, "electrical supply", \textcolor{red}{"workshop"}, \textcolor{red}{"physio and rehabilitation"}, "living dining", \textcolor{red}{"not defined"}, "shaft", \textcolor{red}{"carpark"}, "corridor", \textcolor{red}{"air"}, \textcolor{red}{"dedicated medical room"}, \textcolor{red}{"office space"}, \textcolor{red}{"water supply"}, \textcolor{red}{"garage"}, \textcolor{red}{"medical room"}, \textcolor{blue}{"elevator"}, "balcony", \textcolor{red}{"sanitary rooms"}, "staircase", \textcolor{red}{"vehicle traffic area"}, \textcolor{red}{"cold storage"}, \textcolor{red}{"meeting room"}, "living room", \textcolor{red}{"factory room"}, \textcolor{red}{"showroom"}, \textcolor{red}{"oil tank"}, \textcolor{red}{"office tech room"}, "bedroom", \textcolor{red}{"foyer"}, "room", \textcolor{red}{"patio"}, \textcolor{red}{"teaching room"}, "elevator facilities", \textcolor{red}{"logistics"}, \textcolor{red}{"garden"}, \textcolor{red}{"canteen"}, \textcolor{red}{"community room"}, "gas",\textcolor{red}{ "operations facilities"}, "storeroom", \textcolor{red}{"lobby"}, \textcolor{red}{"shelter"}, \textcolor{red}{"cloakroom"}, \textcolor{red}{"technical area"}, "dining", \textcolor{red}{"warehouse"}, "basement compartment", \textcolor{red}{"loggia"}, \textcolor{red}{"reception room"}, "bathroom", "basement", \textcolor{red}{"common kitchen"}, \textcolor{red}{"pram and bike storage room"}, "bike storage", \textcolor{red}{"break room"}, "house technics facilities", "lightwell", \textcolor{red}{"counter room"}, "transport shaft", \textcolor{red}{"wash and dry room"}, "terrace", \textcolor{red}{"arcade"}, \textcolor{red}{"waiting room"}, "void", "heating", \textcolor{blue}{"kitchen"}, \textcolor{red}{"sports rooms"}, \textcolor{red}{"pram"}, "kitchen dining", \textcolor{red}{"archive"}. 
\end{itemize}

Here, the \textcolor{blue}{blue} indicates which subtype category names are shared between the "feature" and "area" types. 

In addition to the type and subtype categories, each entity contains metadata about the relation to other entities. 
This relation is resembled by the nested positioning of the entities across different sites, buildings, floors, apartments, and units. 
The most high-level positional identifier (ID) is the \textit{site} ID which tells you on which site the entity is located. 
A site could, for instance, be a set of buildings in the same neighborhood (in which the different buildings are likely to share similar characteristics).
Second is the \textit{building} ID which tells you in which building the entity is located. 
Third is the \textit{plan} ID which corresponds to the floor plan layout prototype that the entity is part of. 
Fourth is the \textit{floor} ID which corresponds to a particular floor at a specific elevation in a building. 
It is noteworthy to mention that different floors can originate from the same plan ID.
Fifth is the \textit{apartment} ID which tells you from which apartment the entity originates. 
It is important to note that the apartment ID is shared across different floors in the case of multi-story apartments \ie~apartments that stretch across multiple levels.
Sixth and final is the \textit{unit} ID which indicates from which apartment the entity originates. 
In contrast to the apartment ID, the unit
ID is different for each floor.

Type, subtype, geometry, site ID, building ID, plan ID, floor ID, apartment ID, and unit ID define -- besides other meta-level information such as elevation -- the columns of the dataframe. 
The geometry is defined as a polygon, formatted as \textit{well-known text} (WKT).

\subsection{Filtering details}

For filtering and cleaning, we follow the steps provided in~\sect{data}. 
Some details that were not mentioned specifically in the main text are provided below:

\begin{itemize}
    \item \textbf{Feature removal.} All entities that are a "feature" (see Sec. \ref{subsec:nesting}) are removed entirely from the dataframe.
    \item \textbf{Residential-only filtering.} All floor plans that contain at least one entity for which the subtype category is not to be found in residential buildings (the subtypes indicated in red in Sec. \ref{subsec:nesting}) are removed from the dataframe. 
\end{itemize}

In addition to the filtering steps above, we remove floor plans that have too many small disconnected parts. 
Specifically, we remove all floor plans that have 2 or more areas that are fully disconnected in the room graph (read: that are "floating" in the floor plan); removing an extra 388 (2.8\%) floor plans. 

\subsection{Image extraction}
The coordinates, $x$ (east-to-west) and $y$ (south-to-north), of the geometries in the dataframe are defined in meters and are usually centered around $(x, y) = (0, 0)$ for a given floor plan. To retain the information of the original coordinates within the image, two extra channels are added on top of the image canvas representing $x$ and $y$. The mappings from $x$ and $y$ to the corresponding pixel locations $x_i$ and $y_i$ (both defined on $[0, 512]$) for a given image size $s$ (assumed to be square) are given by:

\begin{align}
    &\begin{aligned}
         x_i =
            \left( x - \textcolor{red}{x_{\text{min}}} + \textcolor{green}{0.5 \left[ \Delta_{yx} \right]_+} \right)
            \cdot
            \textcolor{blue}{\frac{s}{\text{max} \left( \Delta_{x}, \Delta_{y} \right)}},
    \end{aligned} \\
    &\begin{aligned}
        y_i  =
            \left( y - \textcolor{red}{y_{\text{min}}} + \textcolor{green}{0.5 \left[ \Delta_{xy} \right]_+} \right)
            \cdot
            \textcolor{blue}{\frac{s}{\text{max} \left( \Delta_{x}, \Delta_{y} \right)}},
    \end{aligned}
\end{align}

\noindent where $\Delta_{x} = x_{\text{max}} - x_{\text{min}}$ is the 'width' of the floor plan, $\Delta_{y} = y_{\text{max}} - y_{\text{min}}$ is the 'height' of the floor plan, $\Delta_{xy} = \Delta_{x} - \Delta_{y}$ and $\Delta_{yx} = - \Delta_{xy}$ are the relative differences between width and height, and $[\cdot]_+ = \text{max} \left( 0, \cdot \right)$. The red part maps all coordinate values above 0; the green part makes sure to put the floor plan in the middle of the square that starts at $(0,0)$ and extends to $(\text{max}(\Delta_{x}, \Delta_{y}), \text{max}(\Delta_{x}, \Delta_{y}))$; the blue part makes sure to scale the square to the image domain.


\subsection{Statistics}

\begin{figure*}[ht]
    \centering
    \includegraphics[width=\textwidth]{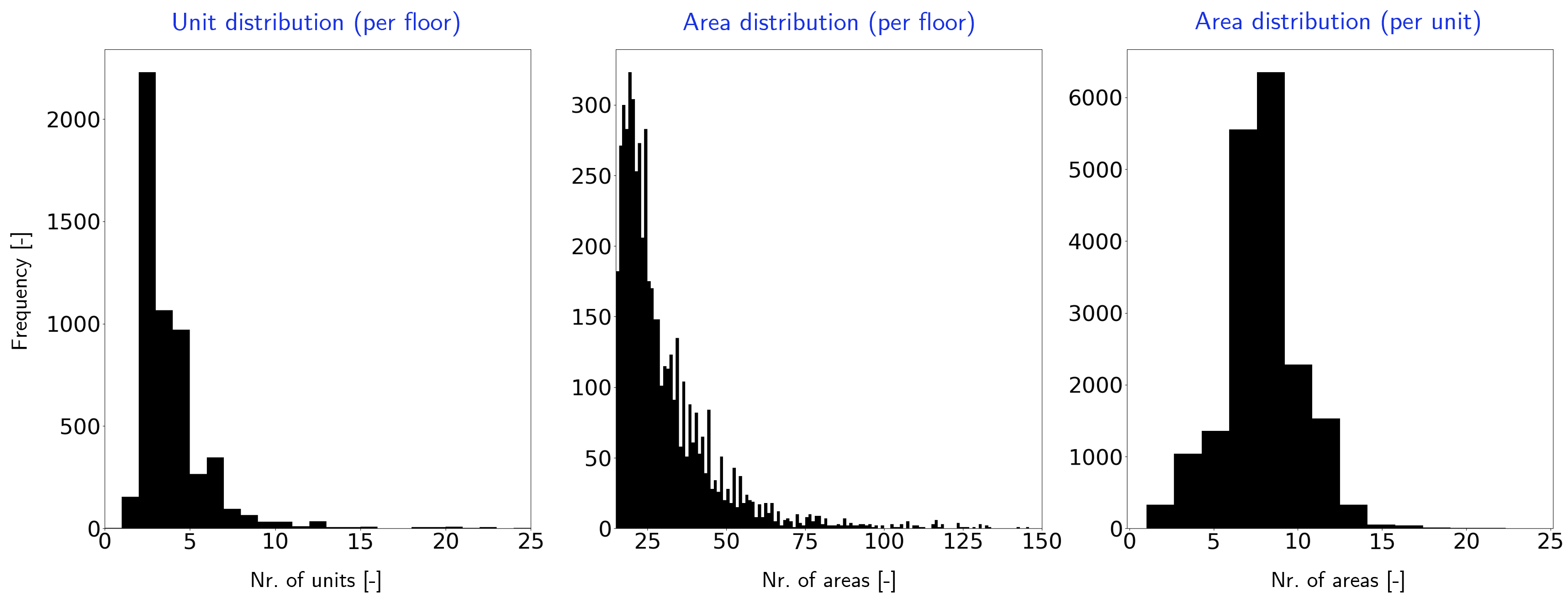}
    \small
    \caption{
        \textbf{Area and unit distributions MSD.} The unit distribution per floor (right), area distribution per floor (middle), and area distribution per unit (right) are plotted as histograms. The x-axis specifies the number of units or areas, and the y-axis specifies the frequency. From the unit distribution plot, it is apparent that MSD comprises mostly floor plans that contain between 2 to 9 units. MSD comprises mostly floor plans that have between 15 and 50 areas, with a peak of around 25. The area distribution per unit is similar to RPLAN~\cite{wu_data-driven_2019} and LIFULL~\cite{lifull_co_ltd_lifull_nodate}, ranging between 3 and 15 areas per unit and a median around 7 areas per unit.
        }
    \label{fig:statistics-msd}
\end{figure*}

With a total of 5.3K+ floor plans, containing 18.9K+ units, and covering 165.3K+ areas, MSD is one of the few publicly available large-scale floor plan datasets\footnote{For details on the sizes of the other floor plan datasets, please refer to \cite{pizarro_automatic_2022}.}. 
\fig{statistics-msd} shows the room and unit distributions for MSD. 

\section{Graph extraction from generated outputs}

In our work, we have set the floor plan generation task to predict only walls and areas -- not doors. 
Therefore, we cannot reliably extract the room graph, $\hat{g}_\mathrm{r}$\footnote{We use $\hat{g}$ to refer to the graph of the predicted floor plan and $g$ to the graph of the ground truth floor plan. Additionally, the subscripts $\mathrm{r}$ and $\mathrm{a}$ refer to 'room' and 'adjacency' graph types.}, from the predicted floor plan because the edge formation critically depends on the door locations. 
Instead, we extract the adjacency graph, $\hat{g}_\mathrm{a}$, from the predicted floor plan in which the edges are formed when geometries are close enough. Note that the set of nodes of $\hat{g}_\mathrm{r}$ and $\hat{g}_\mathrm{a}$ are equivalent, and that the set of edges can be different.

For the graph extraction strategy, we use an algorithmic approach to extract $\hat{g}_\mathrm{a}$.
Specifically, we assign an edge between a pair of nodes if and only if the minimum distance between the areas of that pair does not exceed a preset maximum distance, which is referred to as the buffer distance.
For an image size of 512, we set the buffer distance to 5.
Note that when two areas are overlapping, the minimum distance is 0, hence an edge is formed between overlapping areas.

\begin{figure}[ht]
    \centering
    \includegraphics[width=0.7\linewidth]{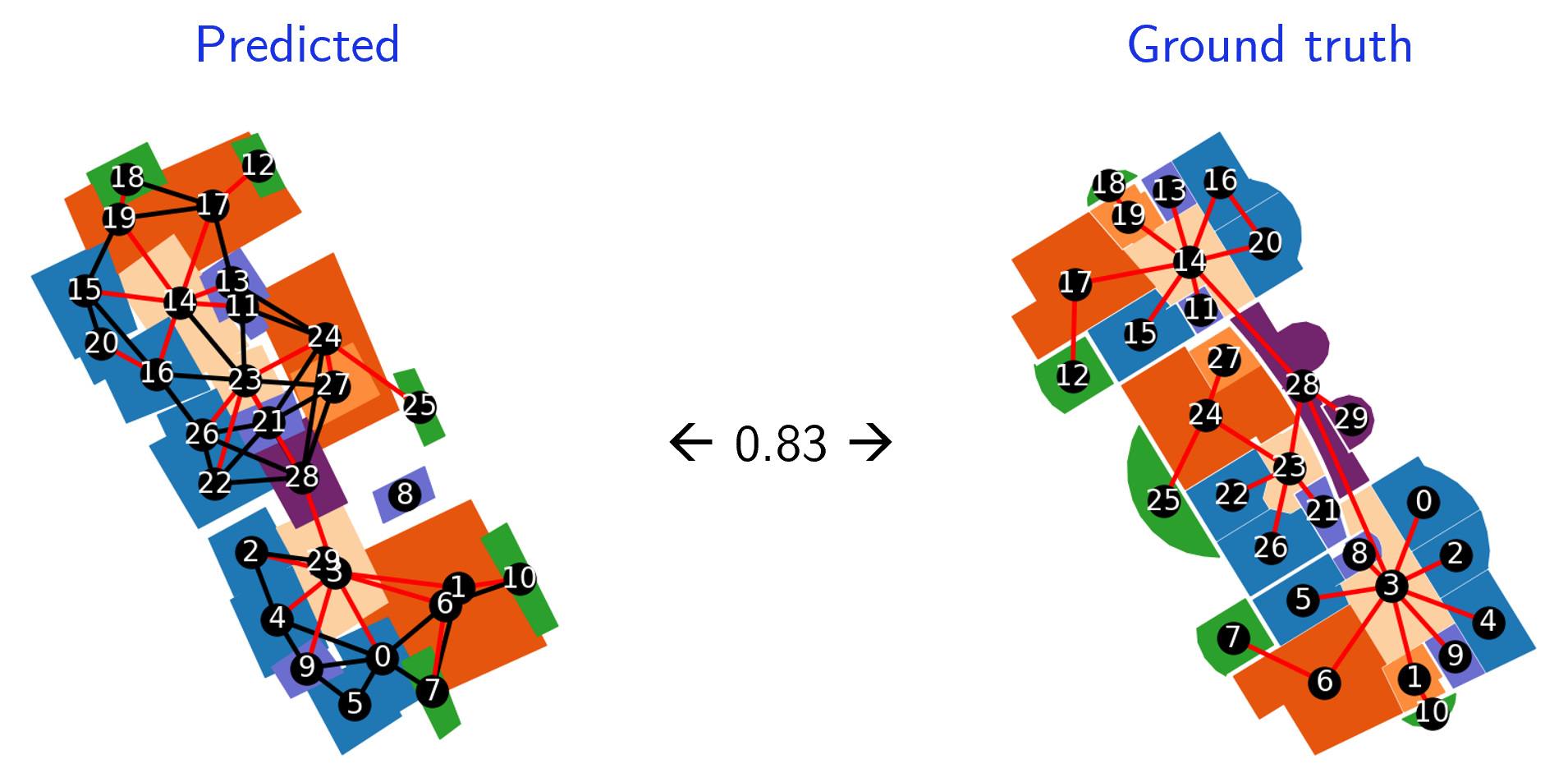}
    \small
    \caption{
        \textbf{Graph compatibility computation.} 
        Left: predicted floor plan including $\hat{g}_\mathrm{a}$. Right: ground truth floor plan including $g_\mathrm{r}$. The set of edges in $g_\mathrm{r}$ that is retained in $\hat{g}_\mathrm{a}$ is colored in red. The node correspondence is made visual by enumerating the nodes graphically. The graph compatibility is computed by dividing the amount of red edges from $\hat{g}_\mathrm{a}$ by the total amount of edges in $g_\mathrm{r}$, equaling $25 / 30 = 0.83$.
        }
    \label{fig:graph-compatibility}
\end{figure}

Access connectivity implies adjacency, but adjacency does not necessarily imply access connectivity, which means that the $g_\mathrm{r}$ is a subgraph of $g_\mathrm{a}$. 
Essentially, the graph compatibility reflects to what extend $g_\mathrm{r}$ is contained in $\hat{g}_\mathrm{a}$, which is done by computing the ratio of the amount of the edges from $g_\mathrm{r}$ that are retained in $\hat{g}_\mathrm{a}$ with respect to the total amount of edges in $\hat{g}_\mathrm{a}$ (see~\eq{compatibility} for details, and ~\fig{graph-compatibility} for a visual elaboration).

For graph-based approaches \eg, MHD~\sect{mhd}, the process of extracting graphs from predicted layouts can also introduce inaccuracies, particularly in cases where the predicted layout encompasses numerous overlapping rooms. 
Nonetheless, our analysis indicates that these instances are infrequent: rooms overlap on average (for MHD) with $4.11 {\scriptstyle \pm 2.25}$ other rooms. 
A typical prediction is given in \fig{graph-compatibility}. 
Therefore, we consider the algorithmic extraction of room edges to be a justifiable method for extracting the room graphs.

\section{Modified HouseDiffusion (MHD)}

\subsection{Node classification with GAT}

We train a GAT model \cite{velickovic_graph_2018} to predict the room graph given the zoning graph. The room and zoning graph are isomorphic, and thus this prediction task can be formulated as a node classification problem. 

The GAT model consists of several stacked graph attention convolutional (GATConv) layers. The input to the model consists of the zoning type for each node, as well as the door type for each edge. The output of the last GATConv is concatenated with the initial node features, and subsequently fed into the final linear layer that maps the concatenated feature vectors to the correct output dimension for predicting the room types. Between each hidden layer, a ReLU is used as an activation function. We use dropout for regularization and use the Adam optimizer. The amount of GATConv layers is set to 5, the hidden sizes of each GATConv to 64, the learning rate to 0.001, and the batch size to 128. Including early stopping, for this setting the validation accuracy is 0.893. 

\subsection{Minimum rotated rectangle approximation (MRR)}

\begin{figure}
    \centering
    \includegraphics[width=0.7\linewidth]{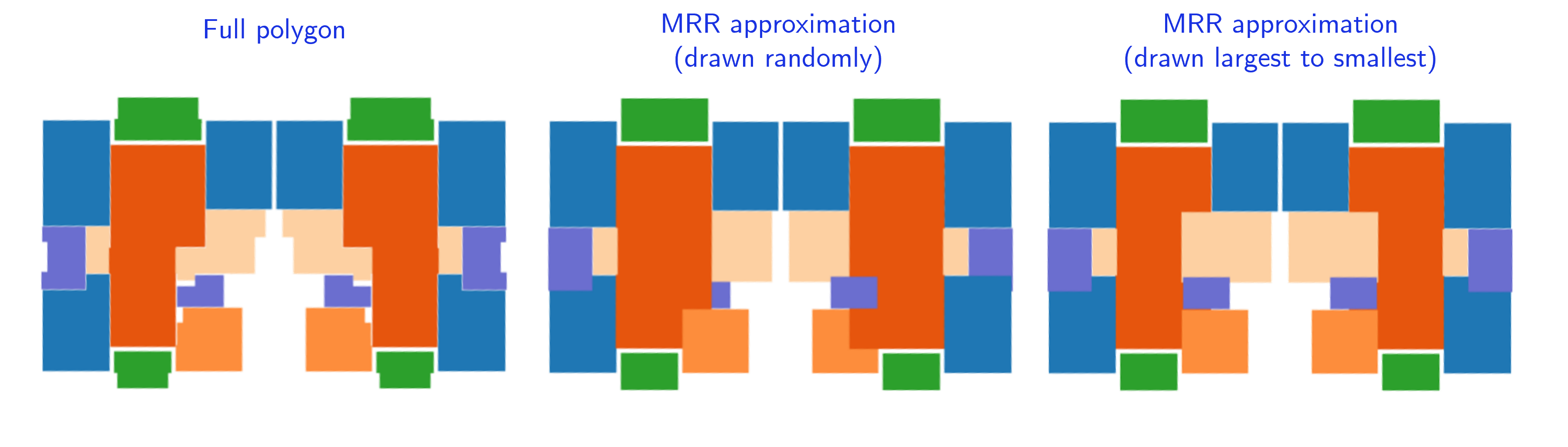}
    \caption{
        \textbf{MRR.} Left: the original floor plan (containing polygons with arbitrarily many corners). Middle: the result of applying MRR, but drawn in random order. Right: the result of applying MRR and drawing the largest to the smallest area.
        }
    \label{fig:mrr-approximation-comparison}
\end{figure}

To be able to represent each area with a fixed number of corners, we propose to take the minimum rotated rectangle of each area. The minimum rotated rectangle of an area is the rotated rectangle that fully encloses the area polygon with minimal area. Approximating the areas of a floor plan by their minimum rotated rectangle (MRR) works best when drawing the area rectangles in order from largest to smallest, such that small areas occlude larger areas (see \cref{fig:mrr-approximation-comparison} for a visual clarification).

\subsection{Diffusion model}
We borrow the model architecture of \textit{HouseDiffusion} (HD)~\cite{shabani_housediffusion_2023}, and extend it to suite MSD. 
The attention layer in the transformer model is modified by adding cross attention between room corners and wall segments. 
Additionally, the relational cross attention (RCA) as defined in HD is modified to incorporate the edge attributes as well. 
RCA is modified because, in contrast to RPLAN in which areas are solely connected by a door, edges in MSD have a "connectivity" attribute representing the type of connectivity being "door", "front door", or "passage". 

We set the batch size to 32 and trained for 300k steps. Other hyperparameters are left the same as in HD implementation.

\subsection{Building structure pre-processing}
To be able to effectively condition MHD on walls, we first convert the binary image of the building structure to a set of straight lines. 
The line elements are extracted from the binary image of the building structure by following the steps below:

\begin{itemize}
    \item \textbf{Morphological thinning.} We start by morphological thinning of the binary image of the building structure. Morphological thinning (see page 671 in \cite{gonzalez_digital_2009}, and an overview of thinning techniques in \cite{lam_thinning_1992}) essentially creates a new binary image in which line thicknesses are reduced to a minimum (ideally one pixel). The resulting binary image is a skeletonized version of the original version.
    \item \textbf{Skeleton network extraction.} From the skeletonized image we extract the skeleton network graph in which nodes represent joints and corners of the skeleton, and edges the geometry of the curves between the nodes. 
    \item \textbf{Simplify skeleton network into set of lines.} The edges of the skeleton network graph, which contain the geometry of the curves between two nodes, are converted to a set of straight lines. 
\end{itemize}

\begin{figure}
    \centering
    \includegraphics[width=0.6\linewidth]{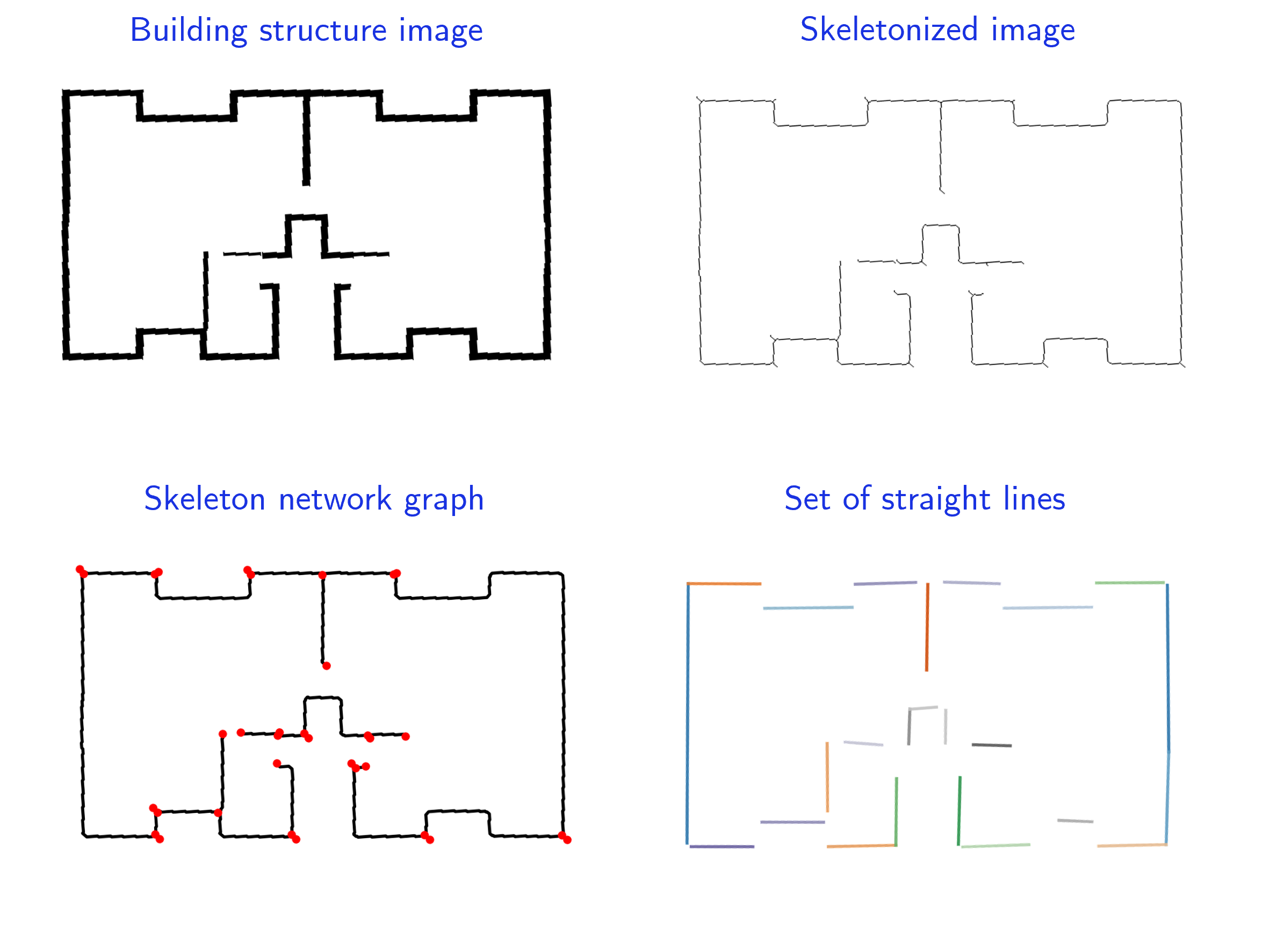}
    \caption{
        \textbf{Line element extraction.} 
        Top-left: building structure as a binary image in which black indicates the necessary structural elements of the building.
        Top-right: a skeletonized building structure in which all the walls are reduced to one-pixel width.
        Bottom-left: a skeleton network graph in which the nodes (joints and corners) are red dots.
        Bottom-right: a simplified set of straight lines approximating the complete building structure as a set of straight-line elements.
        }
    \label{fig:line-element-extraction}
\end{figure}

\cref{fig:line-element-extraction} visualizes the processing steps of the line extraction algorithm.

\subsection{Wall-cross attention (WCA)}
Each wall element $w_i$ extracted by the line extraction algorithm is a vector that represents the start and end points of the line. Similar to HD, we augment $w_i$ by uniformly sampling 7 points between the start and end points. Equivalent to the corner embeddings in HD, a single-layer MLP embeds the 4-D $w_i$ into a 512-D embedding vector: $\hat{w}_i = \mathrm{MLP}_{\mathrm{w}}(\mathrm{AUW}(w_i))$, in which $\mathrm{AUW}$ is the sampling function (similarly named as in HD). Note that the wall elements do not get updated during the denoising process.

\begin{figure}
    \centering
    \includegraphics[width=\linewidth]{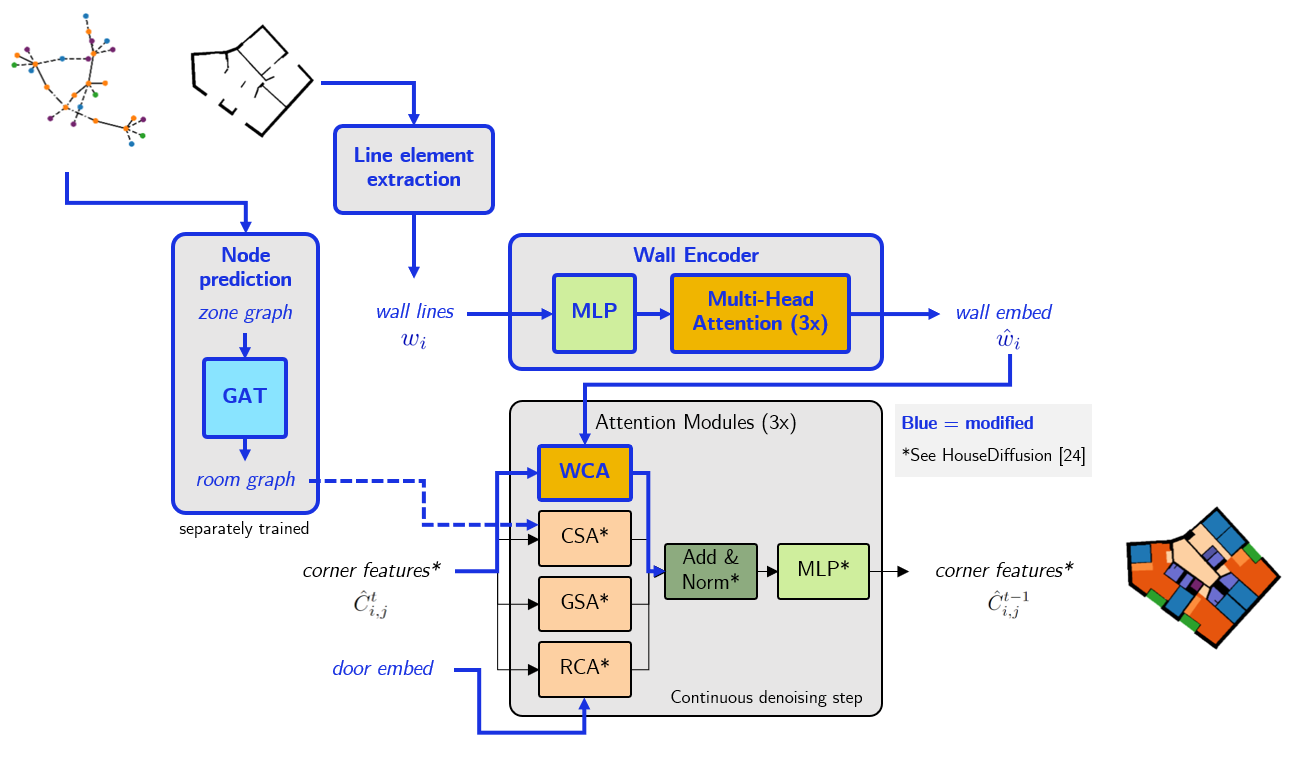}
    \caption{
        \textbf{Modified HouseDiffusion (MHD)}. A wall encoder is used to map the pre-processed building structure into corresponding wall embeddings. MHD expands HD~\cite{shabani_housediffusion_2023} by introducing an extra attention module (WCA) between the wall embeddings from and corner features of the rooms. A GAT is separately trained to predict the room types from the zoning types, which are used to "color" the full layout.
        }
    \label{fig:method-mhd}
\end{figure}

The wall embeddings are used as additional input to MHD. 
In the original model, the attention layer consists of three types of masked attention with room corner embeddings. 
Specifically, the attention module in MHD is modified by adding an extra cross-attention operation between all room corner embeddings and wall embeddings, referred to as wall cross attention (WCA). 
The room corners are used as a query, and the wall embeddings as keys and values. 
All attention operations in the attention layer are summed together.
~\fig{method-mhd} provides a zoomed-in version of~\fig{methods}.

\subsection{RCA with door type embedding}

We modify the RCA module to discern between different connectivity types. Standard doors, passages, and front doors are each assigned a unique learned embedding. The RCA attention is applied separately for each door type, with the attention mask modified to only act on room corners connected by the specified type. On each application, the room corner embeddings are modified when used as keys and values by summing with the embedding of the door type.

\section{Graph-informed U-Net}

\subsection{Visual explanation of model.}

\begin{figure*}[ht]
    \centering
    \includegraphics[width=0.9\textwidth]{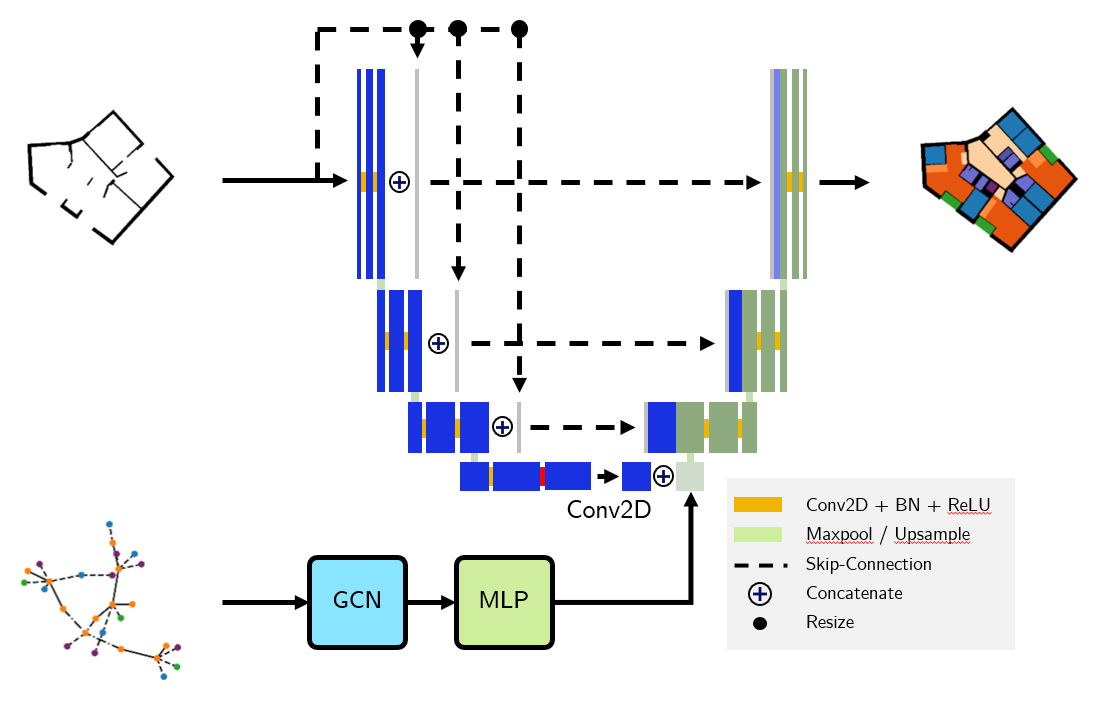}
    \caption{
        \textbf{Graph-informed U-Net (UN).} UN takes the building structure (image) as input to the U-Net. The U-Net is composed of an encoder and decoder using the conventional up- and down-sampling 2D convolutions, resp., and includes skip connections between the encoder and decoder feature maps at equivalent feature map scales. A GCN is used to map the zoning graph to a feature vector which is concatenated to the latent space of the U-Net.  
        }
    \label{fig:un-method}
\end{figure*}

\fig{un-method} shows the architecture of the U-Net model coupled with the GCN. While the U-Net learns a representation for the building structure, the GCN learns a representation for the zoning graph. The two representations are concatenated and simultaneously upsampled by the decoder of the U-Net, outputting the floor plan as a segmented image with the same resolution as the building structure. 

\subsection{Boundary pre-processing}
Initially, the building structure's binary image consists purely of the structural necessary components: black ("0") for structure and white ("1") for non-structure. To better guide the model, we use Segment Anything \cite{kirillov_segment_2023} to predict the interior and exterior of the floor plans and explicitly input that information as well. Before we use Segment Anything, the binary images is substantially padded with extra pixels (white pixels). The padding ensures that the segmentation algorithm can reliably infer the largest area as the exterior, even in cases where the building structure is not completely closed. Once the masks are created, the largest mask is selected as background. The pre-processed image contains the following channels:

\begin{enumerate}
    \item \textbf{"In-wall-out".} This channel marks the interior of the building as '1', the boundaries as '0.5', and the exterior as '0'.
    \item \textbf{"In-out".} This channel marks the interior of the building as '1' and the exterior as '0', focusing on distinguishing between the interior and exterior spaces without structural details.
    \item \textbf{"Raw-boundary".} This channel contains the original building structure.
\end{enumerate}

\subsection{Model, training, and evaluation details}
The encoder of the U-Net comprises four convolutional layers, each with layers that double the channel dimensions from 64 to 512 (64 $\rightarrow$ 128 $\rightarrow$ 256 $\rightarrow$ 512). The convolutional layers all consist of (in order): 3x3 convolution, batch norm, ReLU, and 2x2 maxpool. The GCN consists of a stack of several graph convolutional (GConv) layers, each with a hidden feature size of 256. Global mean pooling is used to compute a graph-level feature vector of size 256. We use the Adam optimizer, and we use the cross-entropy loss.

We found the following optimal settings during training: the amount of GConv layers is 2, a learning rate equal to 0.001, the batch size is 16, and the hidden sizes of each GConv layer are 256.

\section{Additional experiments}

\subsection{Extra baselines: HouseGAN++ and FLNet}
We also ran and evaluated \textit{FLNet}~\cite{upadhyay_flnet_2022} and \textit{HouseGAN++}~\cite{nauata_house-gan_2021} (See~\tab{baselines}).
Both required re-purposing to make them applicable to the task we set. 
All hyperparameters, besides those stated in~\tab{baselines}, are equivalent to those in the original publications.
These additional methods do not perform well, demonstrating the need for our proposed dataset with more realistic building complexes.

\begin{table}[h]
    \caption{
    \small
    \textbf{HouseGAN++}: 128 x 128 masks, 388k steps, learn. rates: 1e-5 generator, 4e-5 discriminator, structural masks as input. \textbf{FLNet}: 128 x 128 masks, 50 epochs. The scores are averaged over all floor plans in the test set. User studies are done for MHD and U-Net: 7 architects, each 50 random IDs. \textbf{Topology}: whether the organization of the spaces makes sense. \textbf{Proportions}: whether the room proportions make sense. Scoring: \{"yes": 1, "unsure": 0.5, "no": 0\}.
    }
    {
    \centering
        \resizebox{1\linewidth}{!}{%
        \begin{tabular}{l|cc|cc}
            
            & MIoU ($\uparrow$) & Compatibility ($\uparrow$) & Topology ($\uparrow$) & Proportions ($\uparrow$)
            \\
            
            \midrule
            
            FLNet
            & 19.3 & n.a. & n.a. & n.a.
            \\
                        
            HouseGAN++ 
            & 11.6 & 64.2 & n.a. & n.a.
            \\

            \midrule

            MHD
            & 21.8 & 76.2 
            & 0.461 $\pm$ 0.138 
            & 0.514 $\pm$ 0.143
            \\

            U-Net
            & 42.4 & n.a. 
            & 0.439 $\pm$ 0.148
            & 0.371 $\pm$ 0.171
            \\
        \end{tabular}
        }
    }
    \label{tab:baselines}
\end{table}

\subsection{MHD on RPLAN}
\begin{figure*}[t]
    \centering
    \includegraphics[width=0.3\textwidth]{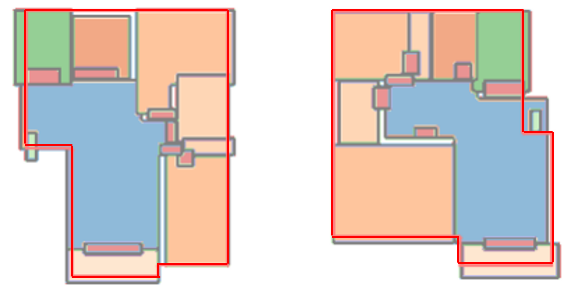}
    \small
    \caption{
        \textbf{MHD on RPLAN.
        }
        Two example predictions of MHD on RPLAN~\cite{wu_data-driven_2019}. The generated layouts follow the boundary reasonably well. The input graphs in both cases are equivalent, showing that the model can cope with a wide variety of differently-shaped boundaries.
        }
    \label{fig:mhd-rplan}
\end{figure*}

Not surprisingly, we successfully trained MHD on RPLAN~\cite{wu_data-driven_2019}, with seemingly similar performance to HD. To train MHD on RPLAN, we extract the boundary of the layouts first. The boundary (as a set of walls) and room graph serve as inputs to MHD. Similar to HD, doors are also predicted (dark red and light green for interior and front doors, resp., in~\fig{mhd-rplan}). Further training details are equivalent to training on MSD. Two typical examples are shown in~\fig{mhd-rplan}.

\subsection{Evaluating complexity}
To better evaluate the complexity, qualitative evaluation (besides the important instrumental measures) will play an essential role.
We are actively researching the evaluation methods for topologically more complex floorplans, and some preliminary results of our study are shown in~\tab{baselines} (right).
Nonetheless, we believe that \textit{both} quantitative as well as qualitative measures play an important role.


\end{document}